\documentclass[letterpaper, 10 pt, conference]{ieeeconf}  

\IEEEoverridecommandlockouts                              
\usepackage[utf8]{inputenc}
\usepackage[T1]{fontenc}

\IEEEoverridecommandlockouts
\usepackage{cite}

\usepackage{amsmath,amssymb,amsfonts}
\usepackage{graphicx}
\usepackage{textcomp}
\usepackage{xcolor}

\usepackage{colortbl}
\usepackage{caption}
\usepackage{subcaption}
\usepackage{algpseudocode}
\usepackage{algorithm}
\algnewcommand\algorithmicforeach{\textbf{for each:}}
\algnewcommand\ForEach{\item[ \algorithmicforeach]}

\graphicspath{ {assets/figures/} }
\usepackage{placeins}
\newcommand{\todo}[1]{}

\newcommand{\etal}{~\textit{et al.}}

\usepackage{tikz}
\newcommand\copyrighttext{%
  \footnotesize \textcopyright 2023 IEEE. Personal use of this material is permitted.
  Permission from IEEE must be obtained for all other uses, in any current or future
  media, including reprinting/republishing this material for advertising or promotional
  purposes, creating new collective works, for resale or redistribution to servers or
  lists, or reuse of any copyrighted component of this work in other works.}
\newcommand\copyrightnotice{%
\begin{tikzpicture}[remember picture,overlay]
\node[anchor=south,yshift=10pt] at (current page.south) {\fbox{\parbox{\dimexpr\textwidth-\fboxsep-\fboxrule\relax}{\copyrighttext}}};
\end{tikzpicture}%
}

\def\BibTeX{{\rm B\kern-.05em{\sc i\kern-.025em b}\kern-.08em
    T\kern-.1667em\lower.7ex\hbox{E}\kern-.125emX}}

\begin{document}

\title{\LARGE \bf\textbf{Hierarchical Graph Neural Networks \\for Proprioceptive 6D Pose Estimation of In-hand Objects}\\

\thanks{$^{1}$Honda Research Institute USA, Inc. {\tt\smallskip \{snehalsubhash\_dikhale, siba, njamali\}@honda-ri.com}}
\thanks{$^{2}$Department of Electrical and Computer Engineering, University of Minnesota. This work was done while he was an intern at Honda Research Institute USA, Inc.
       {\tt\small rezaz003@umn.edu}}

}

\author{Alireza Rezazadeh$^{1, 2}$, 
Snehal Dikhale$^{1}$, 
Soshi Iba$^{1}$ 
and Nawid Jamali$^{1}$
}

\maketitle
\copyrightnotice

\begin{abstract}

Robotic manipulation, in particular in-hand object manipulation, often requires an accurate estimate of the object's 6D pose. To improve the accuracy of the estimated pose, state-of-the-art approaches in 6D object pose estimation use observational data from one or more modalities, e.g., RGB images, depth, and tactile readings. 
However, existing approaches make limited use of the underlying geometric structure of the object captured by these modalities, thereby, increasing their reliance on visual features. 
This results in poor performance when presented with objects that lack such visual features or when visual features are simply occluded.
Furthermore, current approaches do not take advantage of the proprioceptive information embedded in the position of the fingers. 
To address these limitations, in this paper: 
(1) we introduce a hierarchical graph neural network architecture for combining multimodal (vision and touch) data %
that allows for a geometrically informed 6D object pose estimation, 
(2) we introduce a hierarchical message passing operation that flows the information within and across modalities to learn a graph-based object representation, and 
(3) we introduce a method that accounts for the proprioceptive information for in-hand object representation. 
We evaluate our model on a diverse subset of objects from the YCB Object and Model Set, and show that our method substantially outperforms existing state-of-the-art work in accuracy and robustness to occlusion. We also deploy our proposed framework on a real robot and qualitatively demonstrate successful transfer to real settings.  

\end{abstract}


\section{Introduction} \label{sec:introduction}
When humans interact with an object, vision and touch provide information for estimating the object's attributes (e.g., shape, position, and orientation). The nervous system frequently integrates sensory information from several sources, 
where the combined estimate is more accurate than each source individually \cite{gepshtein2005combination, alais2004ventriloquist, ernst2002humans}. For instance, when attempting to grab the handle of a coffee cup, we see and touch the handle. We also see and feel where our hands are based on the proprioception information signaled through muscles and joints \cite{van1999integration, van2002feeling}. The combination of visual, haptic, and proprioceptive feedback allows us to accurately localize the coffee cup and the orientation of its handle. 

State-of-the-art object pose estimation models often use RGB-D observational data to compute an object's 6D pose \cite{wang2019densefusion, wang20206, song2020hybridpose}.  In robotic applications, recent work on multimodal manipulation has shown significant improvements when incorporating tactile data into object reasoning tasks such as object 3D reconstruction \cite{tahoun2021visual}, shape completion \cite{rustler2022active, watkins2019multi}, and pose estimation \cite{dikhale2022visuotactile, bauza2020tactile}. However, these methods make limited use of the underlying geometric structure and disregard proprioceptive information embedded in the position of fingers.

\begin{figure}[t]
    \centering
    \includegraphics[width=\linewidth]{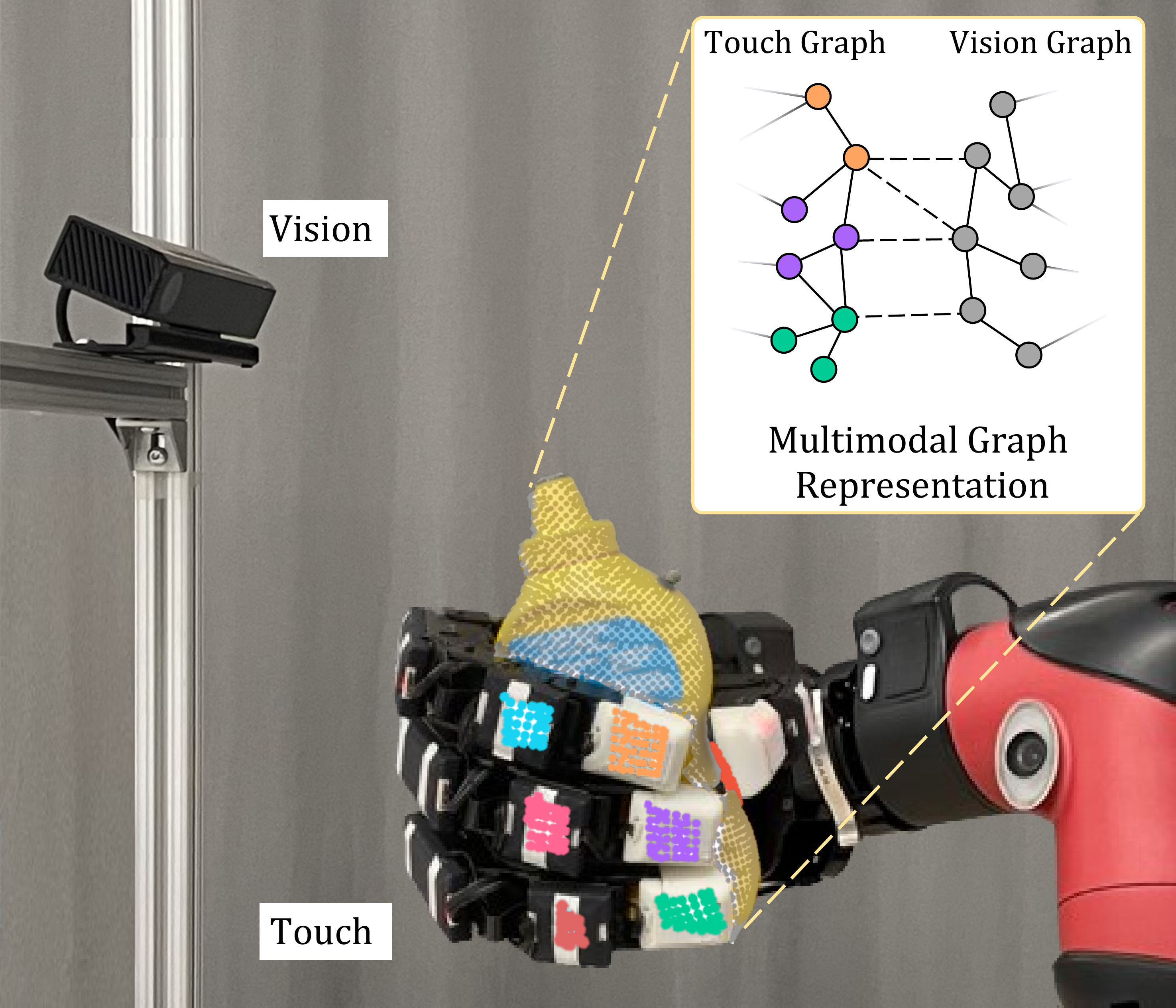}
    \caption{{Our model estimates the 6D pose of an in-hand object by learning a multimodal graph representation from sensor observations. The \textit{Vision Graph} is constructed from the RGB and the depth point cloud shown in gray color, similarly, the \textit{Touch Graph} is constructed from contact points shown in distinct colors for each tactile sensor. These two graphs are also interconnected (dotted lines) to allow the information to flow within and across modalities.}}
    \label{fig:idea}
\end{figure}

Depth information is typically included in 6D pose estimation models as an additional input to a convolutional architecture (CNN) \cite{wang2019densefusion, wang20206, song2020hybridpose}. 
However, CNN architectures do not fully exploit the inherent 3D geometric structure in the depth data. Recently, graph representations have shown promising results in 
capturing the geometric structure of depth point clouds \cite{shi2020point, qi20173d, bi2019graph}.  
Despite the advantage of graph representation for shape completion \cite{rustler2022active, watkins2019multi} and grasp stability prediction \cite{garcia2019tactilegcn}, most existing works on multimodal pose estimation have focused only on CNN architectures \cite{dikhale2022visuotactile, bauza2020tactile}. CNN-based pose estimator models depend predominantly on visual features and underutilize geometric information, which results in significant performance degradation under occlusion.

Earlier work on deep visuomotor policy learning \cite{zhu2018reinforcement, levine2016end} used robot configuration as a complementary input to their model. However, incorporating proprioceptive information such as finger position for in-hand objects has not been explored in the literature. Providing proprioceptive information to the model potentially narrows down the object pose solution space by implicitly filtering out implausible poses that conflict with the hand configuration.

{In this work, we %
introduce a novel framework for in-hand object pose estimation that combines visual and tactile sensor readings and accounts for proprioceptive information. 
We propose learning a multimodal graph representation of an in-hand object on two modality levels: %
a vision graph and a touch graph (see Fig.~\ref{fig:idea}). 
The Vision Graph encodes RGB-D observations while the Touch Graph encodes tactile readings and registers each tactile reading with its corresponding location on the robot's fingers---encapsulating the proprioceptive signal.
We introduce a hierarchical message passing operation that updates the multimodal graph representation by flowing the information within and across the modality graphs. 
The updated multimodal graph representation can then be used to decode an accurate 6D pose of the object.}

Our key contributions are: (1) the proposal of a hierarchical graph neural network architecture for combining vision and touch observations, (2) the introduction of a hierarchical message passing operation to learn a graph-based object representation on multiple modality levels simultaneously through multiple rounds of inter- and intra-modality message passing, and (3) the use of proprioceptive information for in-hand 6D object pose estimation.

\section{Related Work} \label{sec:background}

\subsubsection*{Object Pose from Vision}
Early learning-based pose estimation methods such as Xiang\etal\cite{xiang2017posecnn} explored using RGB input to a convolutional model to directly predict the 6D object pose. Recent work has shown promising improvements by using RGB-D  observations~\cite{wang2019densefusion, wang20206, song2020hybridpose}. Li\etal\cite{li2018unified} included depth data as an additional channel to a convolutional pose estimator. Wang\etal\cite{wang2019densefusion} proposed fusing RGB and depth information at pixel-level to estimate the 6D pose of objects \cite{xiang2017posecnn}. Song\etal\cite{song2020hybridpose} use a hybrid representation based on keypoints to improve pose estimation based on symmetries in the objects. Although RGB-D pose estimation methods are well developed, their performance degrades significantly under occlusion.

\subsubsection*{Object Reasoning from Vision and Touch}
In interactive robotic settings, such as in-hand object manipulation, multi-modal observational data such as vision and touch are available to reason about the characteristics of the object \cite{watkins2019multi, tahoun2021visual, rustler2022active, bauza2020tactile, li2019connecting}. Watkins-Valls\etal\cite{watkins2019multi} and Tahoun\etal\cite{tahoun2021visual} used tactile sensors along with visual RGB-D information to reconstruct 3D geometries of objects under occlusion. Rustler\etal\cite{rustler2022active} also showed 3D shape completion significantly enhances from active re-grasping of the object at the occluded areas. In a more relevant application to our work, Dikhale\etal\cite{dikhale2022visuotactile} proposed a CNN-based framework for 6D pose estimation to combine RGB-D and tactile contact points and showed major improvements in the pose estimation of occluded objects. 

\subsubsection*{Graph-Based Object Reasoning}
Recently graph neural networks have been applied to depth point clouds to process geometric information present in such data. A graph network \cite{scarselli2008graph, battaglia2018relational} is a mapping from an input graph to an updated output graph with updated attributes. Graph neural networks update the graph by learning to perform message passing operations in the graph to propagate information between nodes through edges \cite{gilmer2017neural, battaglia2018relational}. Si\etal\cite{shi2020point} proposed the use of graph representation for object detection based on depth data. Each node in the graph represents a point in the depth point cloud and nodes are connected through edges based on spatial proximity. {For 6D pose estimation, Zhou\etal\cite{zhou2021pr} employed graph convolution \cite{chen2020simple} on RGB-D data and showed improvements over methods with no graph representation such as Wang\etal\cite{wang2019densefusion}.} 

The literature suggests a significant advantage of graph networks for object reasoning based on RGB-D, however, incorporating tactile data with a graph-based model is an underexplored topic. A recent work by Garcia-Garcia\etal\cite{garcia2019tactilegcn} performed grasp stability binary classification by building a graph for pressure readings of tactile sensors. In their method, each tactile sensor is modeled as a separate graph where nodes represent taxels. In comparison, our method aims to estimate the 6D pose of an object by learning a multimodal graph representation that jointly describes the vision and all tactile readings. This allows for capturing the geometric information of each modality as well as their relative geometrical information. We show that our proposed framework reliably estimates an accurate 6D pose and is robust to occlusions.

\section{Problem Definition} \label{sec:problem-definition}
For an object instance within a category (e.g., bottle, can, box, etc.), we refer to category-level 6D pose estimation as the task of predicting the 6D pose $p\in SE(3)$ of the object in the scene with respect to the camera coordinate frame. The 6D pose $p=[R|t]$ consists of a rotation $R \in SO(3)$ and a translation $t \in \mathbb{R}^3$.
We assume access to observational data in two modalities: 
\begin{enumerate}
\item \textit{Vision}: a single-view RGB-D image of the scene that contains RGB scene information $I \in \mathbb{R}^{h \times w \times 3}$ and point cloud of the segmented object $X_V \in \mathbb{R}^{3 \times N_v}$ from the depth channel. 
\item \textit{Touch}: a set of contact points $X_T$ from $N_s$ individual tactile sensors in the form of object surface point clouds $X_T = \{ X_{T_i} \in \mathbb{R}^{3 \times N_{t, i}} \ \textrm{for} \ i=1 \dots N_s \}$. We also assume access to the location of all tactile sensors $X_S \in \mathbb{R}^{3 \times N_s}$ with respect to the camera frame.
\end{enumerate}

\begin{figure*}[t]
    \centering
    \includegraphics[width=\textwidth]{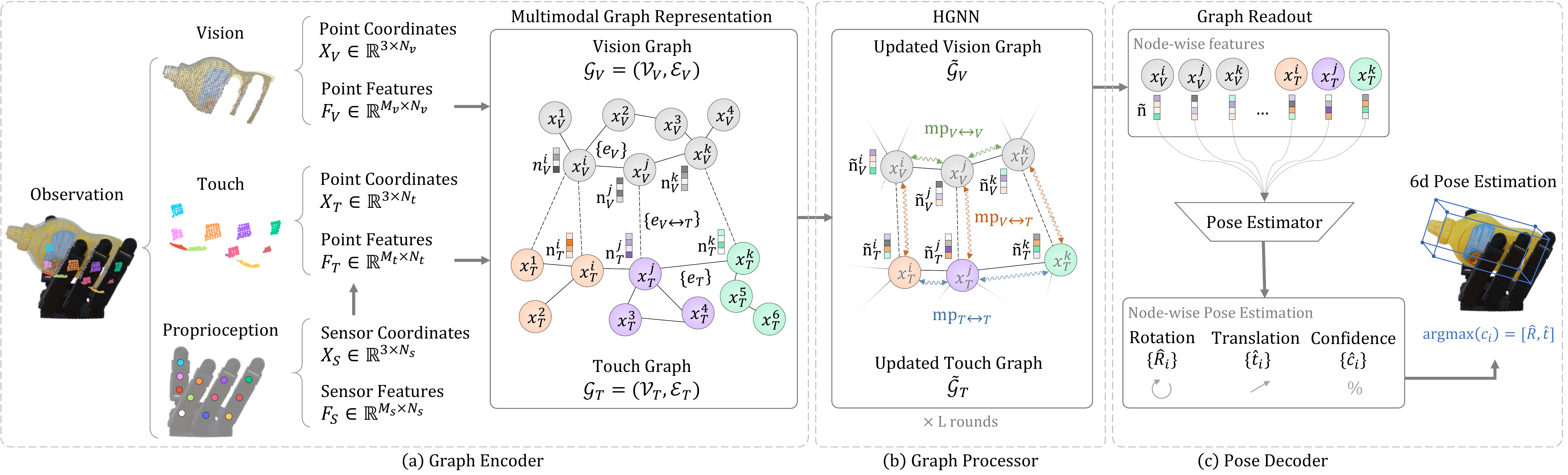}
    \caption{Approach Overview: Our framework estimates the 6D pose of an in-hand object with an encoder-processor-decoder architecture. (a) The graph encoder maps the observation from vision (RGB and depth point cloud) and touch (contact point clouds) to a multimodal graph representation {while accounting for the proprioceptive information from the hand}. (b) The graph processor (HGNN) performs multiple rounds of hierarchical intra-modality, and inter-modality message passing to update the graph. (c) The pose decoder extracts a readout from the updated graph and estimates the object's 6D pose.}
    \label{fig:network-architecture}
\end{figure*}

\section{Model} \label{sec:model}

We propose \textit{Hierarchical Graph Neural Network} (HGNN), a graph neural network framework with an encoder-processor-decoder architecture \cite{battaglia2018relational, sanchez2020learning, pfaff2020learning} to estimate the 6D pose of in-hand objects by combining vision and touch modalities {and accounting for proprioception (Fig.~\ref{fig:network-architecture}).}
 
We introduce learning a hierarchical graph representation of objects on two modality levels: a vision graph, $\mathcal{G_V} = (\mathcal{V}_V, \mathcal{E}_V)$ that encodes observations from vision, and a touch graph, $\mathcal{G_T} = (\mathcal{V}_T, \mathcal{E}_T)$ that encodes touch observations and also encapsulates a proprioceptive signal that registers each tactile reading with the location of its corresponding sensor. These two graphs are interconnected to enable cross-modality information exchange.

\subsection{Model Architecture} \label{sec:model-hgnn}

Our network architecture, as shown in Figure~\ref{fig:network-architecture}, can be divided into three parts: (1) a graph encoder that constructs a hierarchical graph representation of the object, (2) a graph processor that updates the graph representation through hierarchical message passing, and (3) a pose decoder that estimates the 6D pose of the object based on a readout obtained from the updated graph nodes.

\subsubsection{The Graph Encoder} \label{sec:model-encoder}
The graph encoder learns a mapping from the observations to a multimodal graph representation of the object (see Fig.~\ref{fig:network-architecture}-a). First, vision and touch point clouds from raw observation are filtered using voxel downsampling: a 3D voxel grid overlays the points and the centroid of each voxel is used to represent the points within that voxel. 
This preprocessing ensures that our solution is independent of the point cloud density. %
After computing the downsampled vision $X_V$ and touch $X_T$ points, the point features ($F_V$, $F_T$) in each modality and the proprioceptive features ($F_S$) are encoded into a graph representation. 

These features ($F_V$, $F_T, F_S$) contain positional and visual information extracted from the observation. The positional information is based on the location of each point in the 3D space. The visual information is an auxiliary feature to fuse additional relevant information from the image. Visual features are obtained from a bounding box cropped around the 2D projection of the 3D points on the RGB image.  {We refer to the visual feature of the points as auxiliary since they only provide a visual context to the model and are not always directly observable.}

\begin{figure}[tp]
    \centering
    \includegraphics[width=0.65\linewidth]{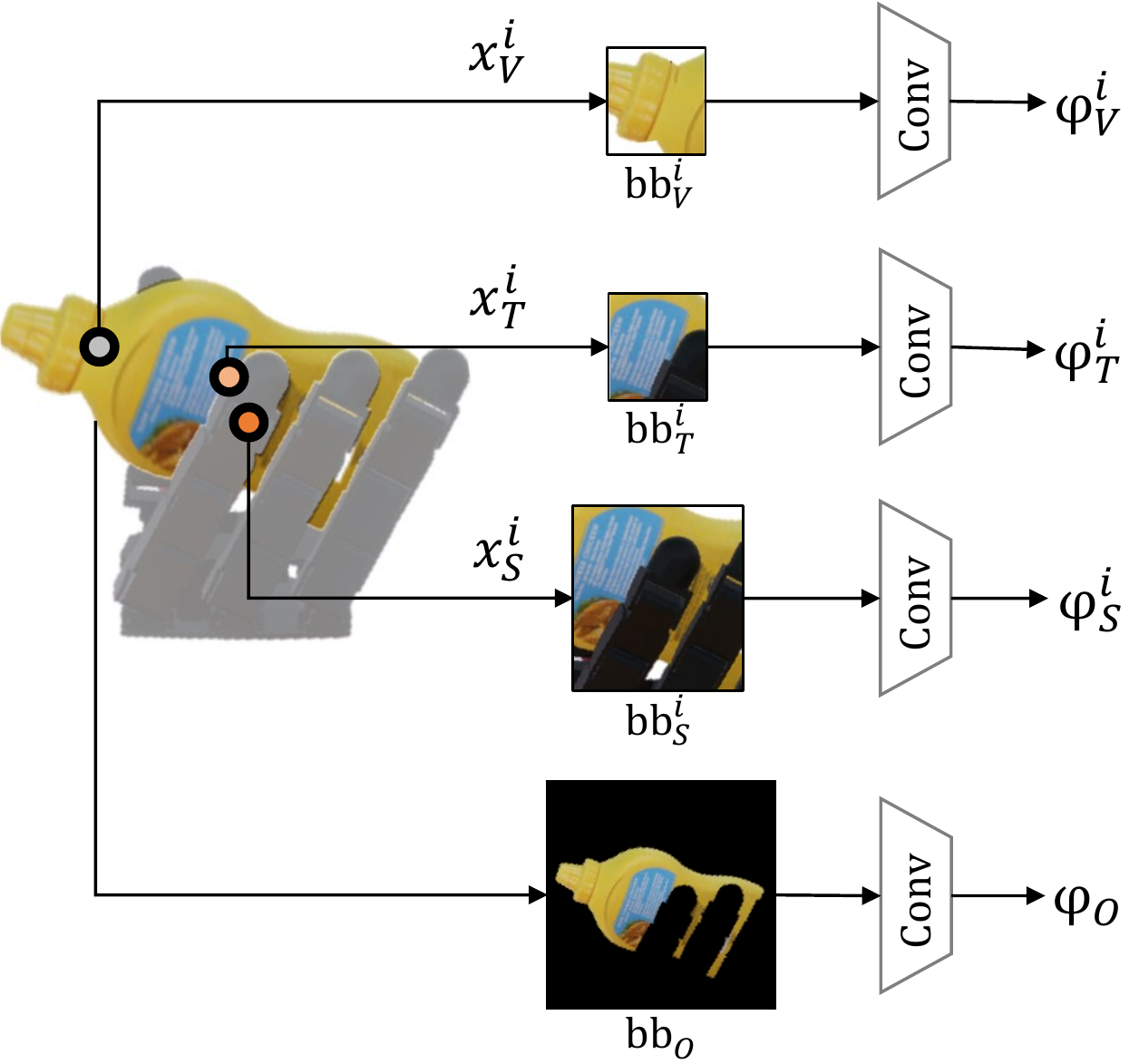}
    \caption{Visual Features: We compute an auxiliary local visual feature vector ($\varphi_V$, $\varphi_T$, $\varphi_S$) for each observed 3D point from vision, touch, and tactile sensors' coordinates ($x_V$, $x_T$, $x_S$) by applying a convolutional encoder to a bounding box cropped around the projection of that point on the RGB image ($\mathrm{bb}_V$, $\mathrm{bb}_T$, $\mathrm{bb}_S$). A global visual feature vector $\varphi_O$ is also computed based on the object segment. }
    \label{fig:network-encoder}
\end{figure}

In the vision graph $\mathcal{G_V}$, each node $\mathbf{n}^i_V \in \mathcal{V}_V$, represents a point in the vision point cloud $X_V$. 
Each vision node embedding is defined as $\mathbf{n}^i_V = [x_V^i, \varphi_V^i, \varphi_O]$ where $x_V^i \in X_V$ denotes the 3D coordinates of the point. 
A local visual feature vector $\varphi_V^i$ is obtained from a convolutional encoder applied to a fixed-size bounding box $\mathrm{bb}_V^i$ around the 2D projection of $x_V^i$ on the observed RGB image. 
A global visual feature vector $\varphi_O$ obtained from a convolutional encoding of the object segment's image $\mathrm{bb}_O$ is also provided in each node embedding (Fig.~\ref{fig:network-encoder}).

The touch graph $\mathcal{G_T}$ carries positional and visual features associated with the tactile readings along with a proprioceptive signal that registers each contact point to its corresponding tactile sensor. Each node embedding in the touch graph is defined as $\mathbf{n}^i_T = [x_T^i, \varphi_T^i, \varphi_O, \boldsymbol\pi_S^i]$ where $x_T^i \in X_T$ is the coordinates of the point, $\varphi_T^i$ is the local visual features from the point's bounding box $\mathrm{bb}_T^i$ on the image, and $\varphi_O$ is the global visual feature of the object (Fig.~\ref{fig:network-encoder}).

For a given node $x_T^i$ in the touch graph, the proprioceptive signal consists of positional and visual features 
 of the corresponding tactile sensor $\boldsymbol\pi_S^i = [x_S^i, \varphi_S^i]$ where $x_S^i \in X_S$ denotes the coordinates of the tactile sensor that observed the point $x_T^i$ and $\varphi_S^i$ is the visual feature obtained from a convolutional encoder applied to bounding box $\mathrm{bb}_S^i$ around the 2D projection of $x_S^i$ on the observed image (Fig.~\ref{fig:network-encoder}). 

Edge embeddings in both graphs $\mathbf{e}^{ij}_V \in \mathcal{E}_V$, and $\mathbf{e}^{ij}_T \in \mathcal{E}_T$ is set by spatial proximity: a node pair $(i, j)$ is connected if $|x_i - x_j|<r$. 
For the connected nodes, the edge embeddings are defined as $|x_i- x_j|$ to capture the relative spatial relation. 
To enable cross-modality information exchange, we also define connectivity between the two graphs through inter-graph edges as $\mathbf{e}^{ij}_{V\leftrightarrow T}$ where each node in a modality graph is connected to the k-nearest neighbor nodes from the other modality graph.

\subsubsection{The Graph Processor} \label{sec:model-processor}
In a graph network $\mathcal{G}$, one round of message passing ($\mathrm{mp}$) updates the graph embeddings by propagating the information in the graph as $\mathcal{G}' \leftarrow \mathrm{mp}(\mathcal{G})$. The message passing operation can be summarized as,
\begin{equation} \label{eq:message-passing}
  \begin{aligned}
    {\mathbf{e}^{ij}}' &\leftarrow f_{\mathrm{e}}(\mathbf{e}^{ij}, \mathbf{n}_i, \mathbf{n}_j)\\
    {\mathbf{n}^{k}}'  &\leftarrow f_{\mathrm{n}}(\mathbf{n}^{k}, \sum_{i \in \mathcal{N}(k)}{\mathbf{e}^{ik}}')
  \end{aligned}
\end{equation}
where an edge specific function $f_{\mathrm{e}}$ updates each node embedding, then for each node the node-specific function $f_{\mathrm{n}}$ aggregates the updated neighboring $\mathcal{N}(k)$ edge information and updates the node embedding. Node- and edge-specific functions are multilayer perceptrons (MLPs). 

We propose Hierarchical Graph Neural Networks (HGNN) with a hierarchical message passing scheme (see Fig.~\ref{fig:network-architecture}-b) where first the information propagates at each modality level and then across modalities. A hierarchical message passing round can be described as,
\begin{equation} \label{eq:hierarchical-message-passing}
  \begin{aligned}
    {\mathcal{G}_V}^{\prime} &\leftarrow \mathrm{mp}_{V\leftrightarrow V}(\mathcal{G}_V)\\
    {\mathcal{G}_T}^{\prime} &\leftarrow \mathrm{mp}_{T\leftrightarrow T}(\mathcal{G}_T)\\
    {\mathcal{G}_V}^{\prime\prime}, {\mathcal{G}_T}^{\prime\prime} &\leftarrow \mathrm{mp}_{V\leftrightarrow T}({\mathcal{G}_V}^{\prime}, {\mathcal{G}_T}^{\prime})\\
  \end{aligned}
\end{equation}
where vision and touch graph nodes communicate within each graph through \textit{intra-modality} message passing ($\mathrm{mp}_{V\leftrightarrow V}$, $\mathrm{mp}_{T\leftrightarrow T}$) and then exchange information through \textit{inter-modality} message passing ($\mathrm{mp}_{V\leftrightarrow T}$). We perform $L$ rounds of hierarchical message passing where each round is applied sequentially to the output of the previous round. After the last message passing round the multimodal graph representation is updated to $\Tilde{\mathcal{G}}$.

\subsubsection{The Pose Decoder} \label{sec:model-decoder}
The updated multimodal graph representation $\Tilde{\mathcal{G}}$ implicitly reflects the object pose information. To extract the object pose, a graph readout is obtained by collecting the updated node embeddings $\Tilde{\mathbf{n}}$ (see Fig.~\ref{fig:network-architecture}-c). We then use a simple MLP pose estimator function to compute a node-wise 6D pose $[\hat{R}_i|\hat{t}_i]$, and a confidence measurement $\hat{c}_i$ for each node-wise pose estimate \cite{wang2019densefusion}. The final pose of the object is set to the node-wise pose estimate with maximum confidence: $\mathrm{argmax}(\hat{c}_i) = [\hat{R}|\hat{t}]$.

\subsection{Model Training} \label{sec:loss}
We use a 3D model of the object to build a loss function to train our framework in an end-to-end manner. First, the object model is voxalized to get a total number of $P$ object surface points $x_p$. Each node-wise estimated pose $[\hat{R}_i|\hat{t}_i]$ is used to transform the object model and then compared with a transformation using the ground-truth object pose $[R_{gt}|t_{gt}]$. A node-wise pose estimation loss is defined as,
\begin{equation} \label{eq:node-loss}
  \begin{aligned}
  L_i^\mathbf{n} = \dfrac{1}{P} \sum_{p = 1 \dots P} || (R_{gt}x_p + t_{gt}) - (\hat{R}_i x_p + \hat{t}_i) ||  
  \end{aligned}
\end{equation}

The total loss is computed as an average node-wise loss $L_i^\mathbf{n}$ weighted by their corresponding confidence estimate $\hat{c}_i$. Top $K$ nodes are pooled to include in the total loss \cite{gao2019graph, knyazev2019understanding, cangea2018towards}. A regularization term is also added for the confidence estimates to achieve a more balanced confidence measurement among all nodes \cite{wang2019densefusion}. The total loss can be summarized as, 
\begin{equation} \label{eq:total-loss}
  \begin{aligned}
  L = \dfrac{1}{K} \sum_{i = 1 \dots K} (L_i^\mathbf{n} \hat{c}_i - \lambda \log (\hat{c}_i))  
  \end{aligned}
\end{equation}

\section{Experiments} \label{sec:experiments}
Our experiments are motivated by the following questions: (1) Does the HGNN model learn to accurately estimate the object pose? (2) How effective is incorporating the proprioceptive information? (3) Does hierarchical message passing improve performance? 

\subsection{Environments} \label{sec:exp-env}
\subsubsection{Synthetic} \label{sec:exp-env-sim}
We use VisuoTactile synthetic dataset from Dikhale\etal\cite{dikhale2022visuotactile} to train our framework. In this dataset, a subset of 11 YCB objects \cite{calli2015ycb} are selected based on their graspability. A total number of $20$K distinct in-hand poses are simulated per object. In particular, Unreal Engine 4.0 \cite{unrealengine} has been used to render photo-realistic observational data of a 6 DoF robot arm with a 4-fingered gripper (Allegro Hand, SimLab Co., Ltd.) equipped with 12 tactile sensors (3 per finger). A main RGB-D camera captures images of the robot holding an object. Each tactile sensor captures object surface contact points in a point cloud format. Each data sample is generated by randomizing the in-hand object pose, the robot fingers configuration, and the robot arm orientation and position. Domain randomization is also applied for the color and pattern of the background and workspace desk.

\subsubsection{Real Robot} \label{sec:exp-env-real}
After training our model on the synthetic data we deploy it on a multi-finger gripper (Allegro Hand, SimLab Co., Ltd.) attached to a Sawyer robot. The gripper has 4 fingers, 16 joints, and 3 tactile sensors (uSkin, XELA Robotics Co., Ltd) on each finger to capture the object's surface contact points (224 taxels in total). An RGB-D camera (Kinect2, Microsoft) is used as the main camera to get RGB-D images. We use real YCB object samples to test the performance of our framework on the real robot (Fig.~\ref{fig:idea} shows the real robot setup).

\subsection{Baselines and Ablations} \label{sec:exp-baseline}

We compare our approach (HGNN) with two baselines:
\vspace{0.125cm}
    \subsubsection{VisuoTactile Fusion (ViTa)} we compare with a recent multimodal category-level 6D object pose estimation Dikhale\etal\cite{dikhale2022visuotactile}. Vision and tactile data are processed in two separate channels. In the vision channel, RGB and depth data are fused at the pixel level. The tactile channel combines the depth data and tactile contact points. The outputs of the two channels are fused using 1D convolutional layers to estimate the 6D pose. 

    \subsubsection{Point Cloud Graph Neural Network (Point-GNN)} we modified a recent graph-based object detection method by Shi\etal\cite{shi2020point} to build a graph-based 6D pose estimation baseline. A graph representation is built where each node is a point in the point cloud. Unlike our hierarchical scheme, Shi\etal\cite{shi2020point} applied message passing across all nodes to update the graph. We use a pose decoder architecture similar to ours (section \ref{sec:model-decoder}) to estimate the object 6D pose.

\vspace{0.25cm}
\setcounter{subsubsection}{0}
We also ablate our model to examine our formulation and single out the contribution of each of its components:
\vspace{0.125cm}

    \subsubsection{Visual Features (HGNN-NoVis)} In this model, we only rely on the 3D coordinates of the observed points (depth, touch, and proprioception) and exclude all auxiliary visual features obtained from convolutional encoders ($\varphi_O$, $\varphi_V$, $\varphi_T$, and $\varphi_S$). 

    \subsubsection{Proprioception (HGNN-NoProp)} In this model, we exclude all proprioceptive information $\boldsymbol\pi_S$ (positional and visual features of the tactile sensors) from the graph representation of the observational data.
        
    \subsubsection{Hierarchical Message Passing (HGNN-NoHrch)} We discard the hierarchical inter- and intra-modality message passing from equation (\ref{eq:hierarchical-message-passing}). Instead, we perform $L=3$ rounds of message passing across all nodes with no hierarchical scheme.

\subsection{Metrics} \label{sec:exp-metrics}
We measure the quality of an estimated 6D pose $\hat{p}=[\hat{R}|\hat{t}]$ using two metrics: position error and angular error. The position error is computed as L2 distance between the estimated $\hat{t}$ and ground truth translation vectors $t_{gt}$. The angular error is defined as $\cos^{-1}(2\langle \hat{q},q_{gt} \rangle^2 -1)$ using the inner product of the estimated $\hat{q}$ and ground truth $q_{gt}$ rotation quaternions \cite{huynh2009metrics}.

\subsection{Implementation Details} \label{sec:impl-details}
All models are trained on the synthetic dataset. The dataset is split into training and testing sets, $80\%$ and $20\%$, respectively. We report the performance of models using the testing data. The bounding box sizes are set to $\mathrm{bb}_V, \mathrm{bb}_T: 8\times8$ pixels and $ \mathrm{bb}_S: 64\times64$ pixels. A total number of $L=3$ rounds of message passing has been performed in equation (\ref{eq:hierarchical-message-passing}). All the concatenated features in embeddings of the vision and touch graphs are mapped to a fixed-size 128-dimensional latent vector using MLPs. We set $\lambda=1.5\text{e-}2$ and $K=128$ in equation (\ref{eq:total-loss}).


\section{Results} \label{sec:results}

\begin{figure}[t]
    \centering
    \includegraphics[width=\linewidth]{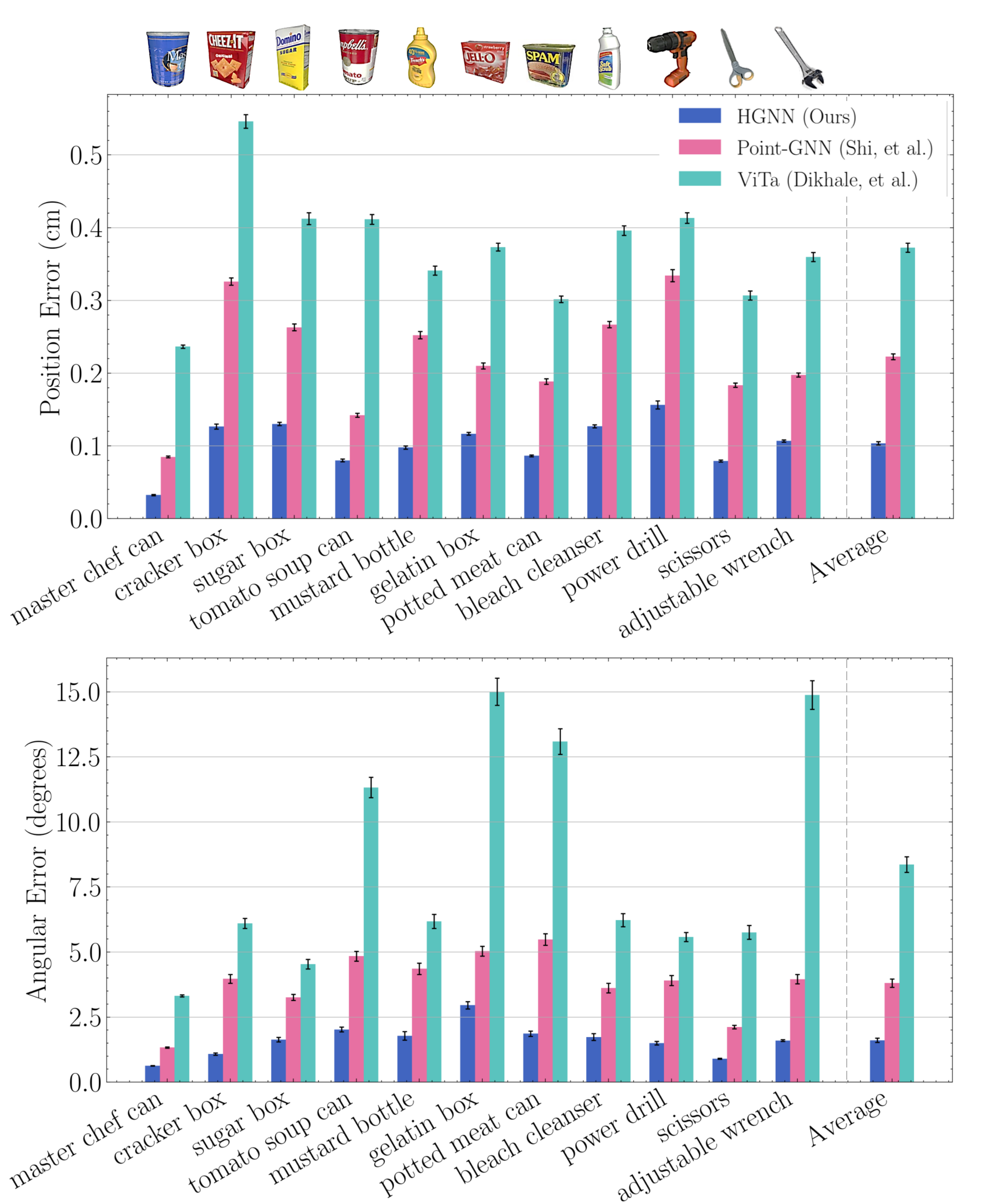}
    \caption{6D pose estimation performance.}
    \label{fig:result-baselines}
\end{figure}
\begin{figure}
    \centering
    \includegraphics[width=\linewidth]{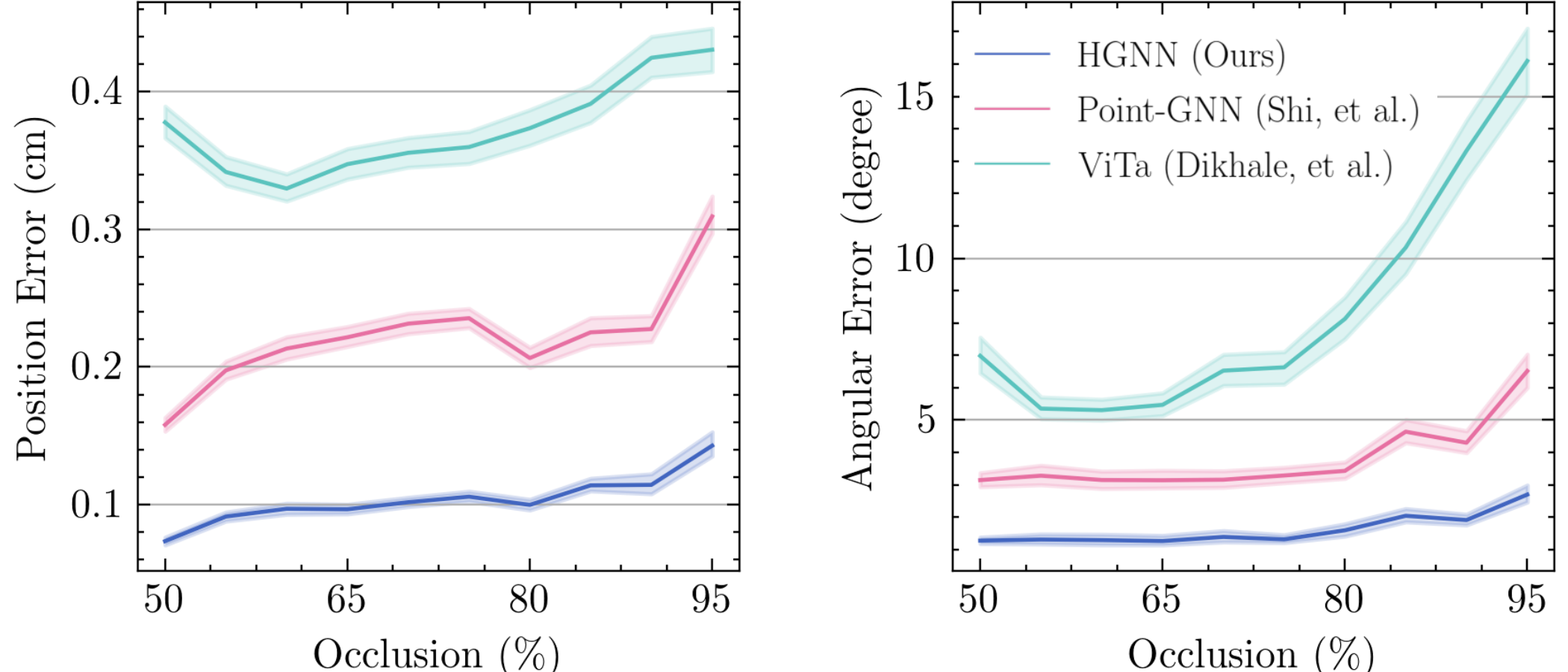}
    \caption{6D pose estimation performance under increasing levels of occlusion.}
    \label{fig:result-occlusion}
\end{figure}
\subsection{Comparison with Baselines} \label{sec:res-baselines}
 
Figure \ref{fig:result-baselines} shows the performance of our model measured in angular (deg) and position (cm) error.
Across all objects, our model \textit{HGNN} consistently outperforms \textit{Point-GNN} and \textit{ViTa} by a large margin. On average, \textit{HGNN} outperforms \textit{Point-GNN}, by $53\%$ ($0.12$ cm) in position and $57\%$ ($2.20 ^{\circ}$) in angular error. The improvement margin is larger in comparison with \textit{ViTa}, by $72\%$ ($0.27$ cm) in position and $80\%$ ($6.7 ^{\circ}$) in angular error.

Another important observation comes from examining the effect of occlusion on the performance of our model. We calculate the occlusion level as the percentage of object model points visible to the main camera. Each point in the model is labeled as visible if there exists a point within a fixed threshold in the main camera's point cloud using the k-nearest neighbor algorithm. The occlusion level is the percentage of model points labeled as not visible. Figure~\ref{fig:result-occlusion} shows the results of this analysis. Our model is effectively more robust to occlusion levels compared to the baseline.

\subsection{Ablations} \label{sec:res-ablate}
 
Table \ref{table:result-ablate} summarizes the ablation experiment results.

\subsubsection{Effects of Proprioceptive Information} \label{sec:res-ablate-prop}

We remove the proprioceptive information from the graph representation in \textit{HGNN-NoProp} and observe a significantly inferior performance compared to \textit{HGNN} ($23\%$ in position and $33\%$ in angular error).

\subsubsection{Effects of Visual Features} \label{sec:res-ablate-vis}

We exclude the auxiliary visual features obtained from bounding boxes and only rely on 3D coordinates of the observed points in \textit{HGNN-NoVis}. This results in a significantly less accurate pose estimation compared to \textit{HGNN} ($18\%$ in position and $15\%$ in angular error) which validates that providing visual context for the points enhances the learned object representation.

\subsubsection{Effects of Hierarchical Message Passing} \label{sec:res-ablate-mp}

We discard the hierarchical message passing and use simple message passing across all modalities in \textit{HGNN-NoHrch}. No hierarchy in message passing results in significantly inferior performance ($15\%$ in position and $70\%$ in angular error). This indicates that a hierarchical scheme (i.e., inter- and intra-modality message passing) has a major advantage for multimodal graph representation learning.

\begin{table}
\centering
    \caption{Ablation Results.}
    \label{table:result-ablate}
    \begin{tabular}{|l|cc|}
    \hline
    Model & Position Error (cm) & Angular Error (degree)\\
    \hline
        HGNN (Ours)   &$\mathbf{0.104 \pm 0.002}$   &$\mathbf{1.609 \pm 0.081}$\\  
        HGNN-NoVis         &$0.127 \pm 0.003$            &$1.907 \pm 0.111$\\
        HGNN-NoProp       &$0.136 \pm 0.002$            &$2.390 \pm 0.122$\\
        HGNN-NoHrch   &$0.154 \pm 0.002$            &$2.749 \pm 0.108$\\
    \hline
    \end{tabular}
\end{table}

\subsubsection{Effect of Tactile Readings} \label{sec:res-ablate-tact}

\begin{figure}[t]
    \centering
    \includegraphics[width=\linewidth]{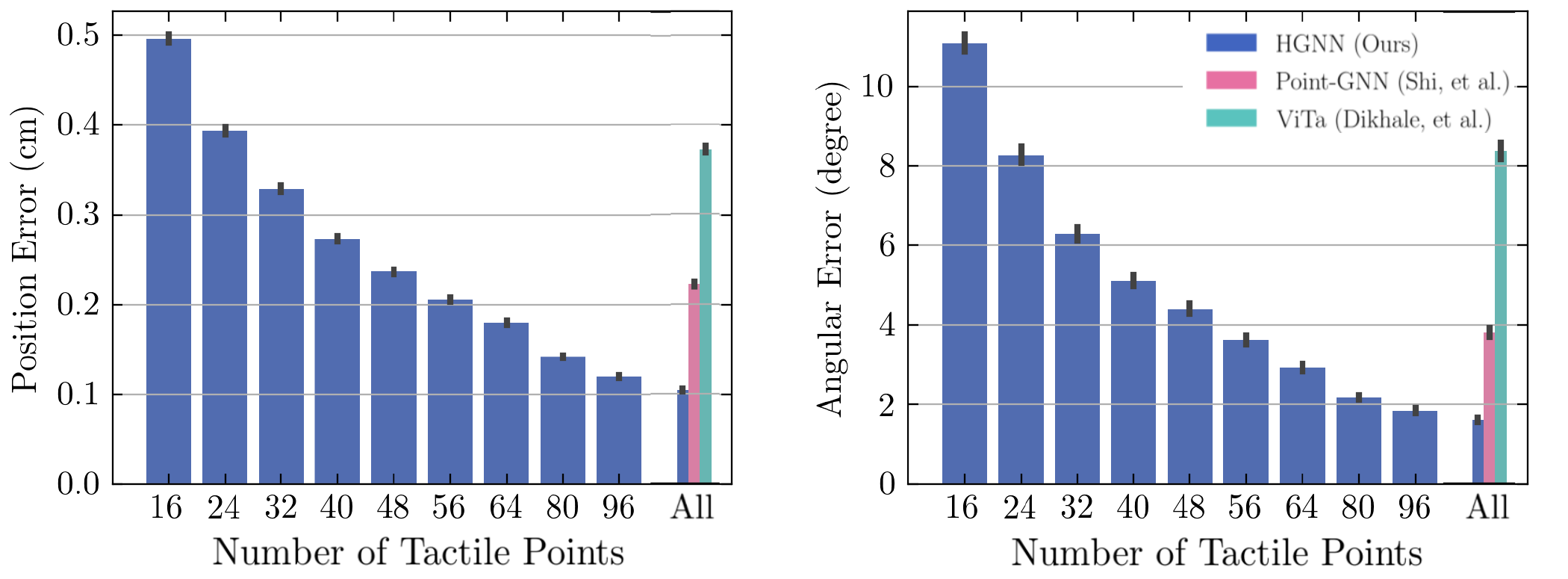}
    \caption{Effect of the varying maximum number of tactile points per sensor on HGNN pose estimation performance.}
    \label{fig:result-vartactile}
\end{figure}

We assess the performance of our model over varying numbers of tactile data. Figure~\ref{fig:result-vartactile} shows the average accuracy of our model as a response to limiting the maximum number of tactile contact points available per sensor. We notice that as the number of points per sensor increases (i.e, the touch observation gets richer), our model increasingly estimates a more accurate pose. This shows the importance of incorporating the touch modality on top of vision to enhance the quality of the estimated pose. 

\subsection{Real Robot} \label{sec:res-realrobot}
\begin{figure}[t!]
    \centering
    \includegraphics[width=0.48\textwidth]{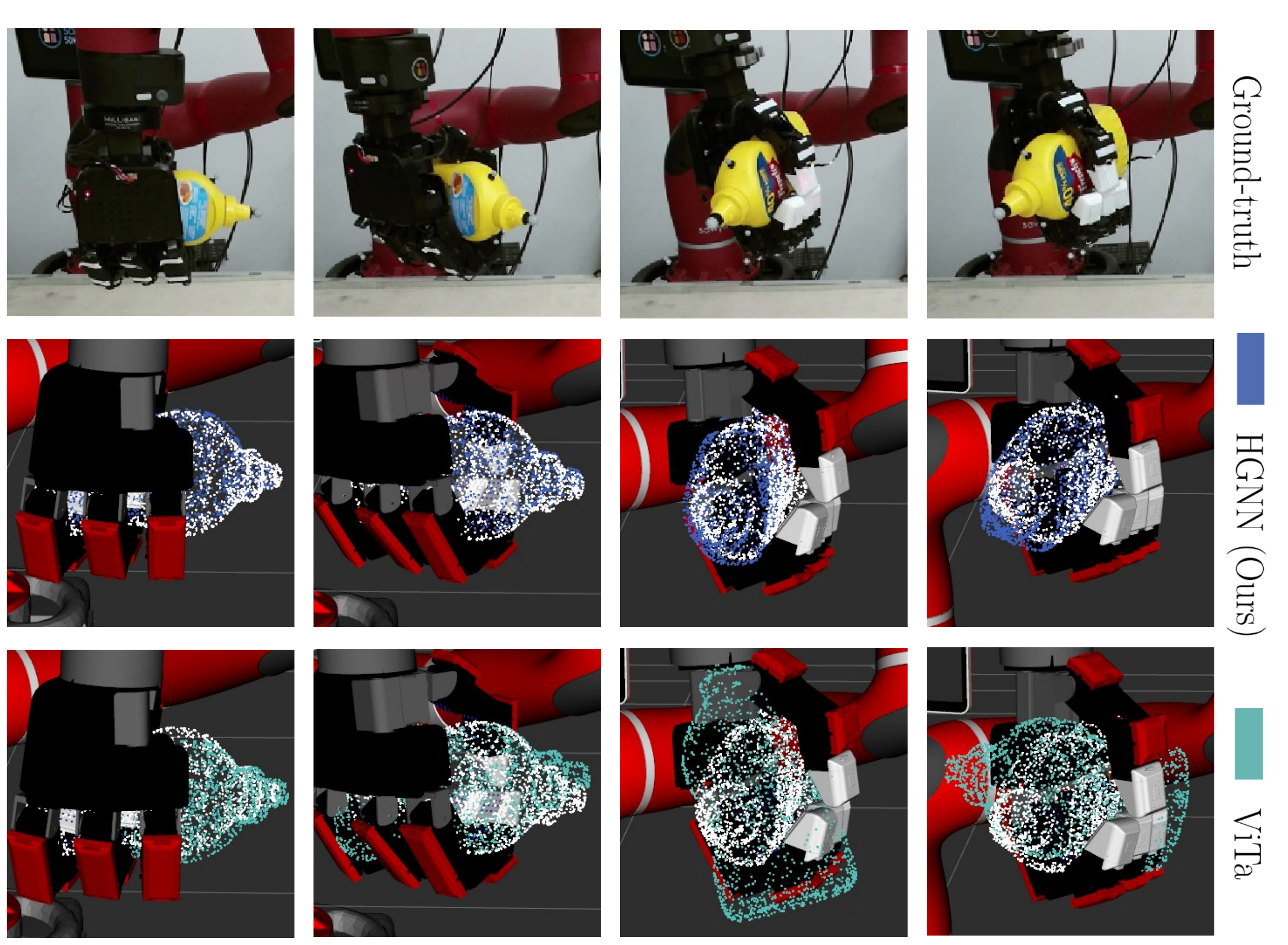}
    \caption{Pose estimation performance under different levels of occlusion for HGNN (blue), ViTa (cyan) compared with the ground-truth (gray).}
    \label{fig:robot_occlusion}
\end{figure}

We deployed our HGNN network and ViTa baseline network simultaneously in real-time on the hardware setup described in section~\ref{sec:exp-env-real}. During deployment, both our network and the baseline receive the same input. Note, that the networks are trained on the synthetic dataset only. Figure~\ref{fig:robot_occlusion} is an example of the pose estimation performance under varying levels of occlusion. We observe that our model (blue), unlike ViTa (cyan), is robust to a wide range of occlusions. The accompanying video provides further demonstration of the results presented in this section.

\section{Conclusion} \label{sec:conclusion}
In this paper, we proposed a graph-based framework for estimating the 6D pose of in-hand objects based on vision and touch observations and accounting for proprioceptive information.
Motivated by human motor behavior, we introduced a novel hierarchical approach for learning a graph representation of the observational data on two modality levels using hierarchical message passing that allows the information to flow within and across modalities.
We showed that our approach accurately estimates the object's pose and is robust to heavy occlusions.
We compared our model to existing work and showed that it achieves state-of-the-art performance on category-level 6D pose estimation of YCB objects.
Moreover, we deployed our model to a real robot and showed successful transfer to pose estimation in the real setting.
One potential limitation in our formulation is that incorporating proprioceptive information leads to learning a pose estimator that is not gripper-agnostic. Although we showed that including proprioception significantly enhances the pose estimation accuracy, we speculate that for optimal performance the model needs to be retrained based on the gripper characteristics (e.g., number of fingers or number of sensors per finger). 
An interesting future direction is incorporating other sensory information such as pressure into our framework. Finally, we hope our general approach inspires future research on multimodal object representation learning.

\bibliographystyle{IEEEtran}
\bibliography{IEEEabrv, ./bibliography/references-all.bib}

\begin{thebibliography}{10}
\providecommand{\url}[1]{#1}
\csname url@samestyle\endcsname
\providecommand{\newblock}{\relax}
\providecommand{\bibinfo}[2]{#2}
\providecommand{\BIBentrySTDinterwordspacing}{\spaceskip=0pt\relax}
\providecommand{\BIBentryALTinterwordstretchfactor}{4}
\providecommand{\BIBentryALTinterwordspacing}{\spaceskip=\fontdimen2\font plus
\BIBentryALTinterwordstretchfactor\fontdimen3\font minus
  \fontdimen4\font\relax}
\providecommand{\BIBforeignlanguage}[2]{{%
\expandafter\ifx\csname l@#1\endcsname\relax
\typeout{** WARNING: IEEEtran.bst: No hyphenation pattern has been}%
\typeout{** loaded for the language `#1'. Using the pattern for}%
\typeout{** the default language instead.}%
\else
\language=\csname l@#1\endcsname
\fi
#2}}
\providecommand{\BIBdecl}{\relax}
\BIBdecl

\bibitem{gepshtein2005combination}
S.~Gepshtein, J.~Burge, M.~O. Ernst, and M.~S. Banks, ``The combination of
  vision and touch depends on spatial proximity,'' \emph{Journal of vision},
  vol.~5, no.~11, pp. 7--7, 2005.

\bibitem{alais2004ventriloquist}
D.~Alais and D.~Burr, ``The ventriloquist effect results from near-optimal
  bimodal integration,'' \emph{Current biology}, vol.~14, no.~3, pp. 257--262,
  2004.

\bibitem{ernst2002humans}
M.~O. Ernst and M.~S. Banks, ``Humans integrate visual and haptic information
  in a statistically optimal fashion,'' \emph{Nature}, vol. 415, no. 6870, pp.
  429--433, 2002.

\bibitem{van1999integration}
R.~J. Van~Beers, A.~C. Sittig, and J.~J. D. v.~d. Gon, ``Integration of
  proprioceptive and visual position-information: An experimentally supported
  model,'' \emph{Journal of neurophysiology}, vol.~81, no.~3, pp. 1355--1364,
  1999.

\bibitem{van2002feeling}
R.~J. van Beers, D.~M. Wolpert, and P.~Haggard, ``When feeling is more
  important than seeing in sensorimotor adaptation,'' \emph{Current biology},
  vol.~12, no.~10, pp. 834--837, 2002.

\bibitem{wang2019densefusion}
C.~Wang, D.~Xu, Y.~Zhu, R.~Mart{\'\i}n-Mart{\'\i}n, C.~Lu, L.~Fei-Fei, and
  S.~Savarese, ``Densefusion: 6d object pose estimation by iterative dense
  fusion,'' in \emph{Proceedings of the IEEE/CVF conference on computer vision
  and pattern recognition}, 2019, pp. 3343--3352.

\bibitem{wang20206}
C.~Wang, R.~Mart{\'\i}n-Mart{\'\i}n, D.~Xu, J.~Lv, C.~Lu, L.~Fei-Fei,
  S.~Savarese, and Y.~Zhu, ``6-pack: Category-level 6d pose tracker with
  anchor-based keypoints,'' in \emph{2020 IEEE International Conference on
  Robotics and Automation (ICRA)}.\hskip 1em plus 0.5em minus 0.4em\relax IEEE,
  2020, pp. 10\,059--10\,066.

\bibitem{song2020hybridpose}
C.~Song, J.~Song, and Q.~Huang, ``Hybridpose: 6d object pose estimation under
  hybrid representations,'' in \emph{Proceedings of the IEEE/CVF conference on
  computer vision and pattern recognition}, 2020, pp. 431--440.

\bibitem{tahoun2021visual}
M.~Tahoun, O.~Tahri, J.~A.~C. Ram{\'o}n, and Y.~Mezouar, ``Visual-tactile
  fusion for 3d objects reconstruction from a single depth view and a single
  gripper touch for robotics tasks,'' in \emph{2021 IEEE/RSJ International
  Conference on Intelligent Robots and Systems (IROS)}.\hskip 1em plus 0.5em
  minus 0.4em\relax IEEE, 2021, pp. 6786--6793.

\bibitem{rustler2022active}
L.~Rustler, J.~Lundell, J.~K. Behrens, V.~Kyrki, and M.~Hoffmann, ``Active
  visuo-haptic object shape completion,'' \emph{IEEE Robotics and Automation
  Letters}, vol.~7, no.~2, pp. 5254--5261, 2022.

\bibitem{watkins2019multi}
D.~Watkins-Valls, J.~Varley, and P.~Allen, ``Multi-modal geometric learning for
  grasping and manipulation,'' in \emph{2019 International conference on
  robotics and automation (ICRA)}.\hskip 1em plus 0.5em minus 0.4em\relax IEEE,
  2019, pp. 7339--7345.

\bibitem{dikhale2022visuotactile}
S.~Dikhale, K.~Patel, D.~Dhingra, I.~Naramura, A.~Hayashi, S.~Iba, and
  N.~Jamali, ``Visuotactile 6d pose estimation of an in-hand object using
  vision and tactile sensor data,'' \emph{IEEE Robotics and Automation
  Letters}, vol.~7, no.~2, pp. 2148--2155, 2022.

\bibitem{bauza2020tactile}
M.~Bauza, E.~Valls, B.~Lim, T.~Sechopoulos, and A.~Rodriguez, ``Tactile object
  pose estimation from the first touch with geometric contact rendering,''
  \emph{arXiv preprint arXiv:2012.05205}, 2020.

\bibitem{shi2020point}
W.~Shi and R.~Rajkumar, ``Point-gnn: Graph neural network for 3d object
  detection in a point cloud,'' in \emph{Proceedings of the IEEE/CVF conference
  on computer vision and pattern recognition}, 2020, pp. 1711--1719.

\bibitem{qi20173d}
X.~Qi, R.~Liao, J.~Jia, S.~Fidler, and R.~Urtasun, ``3d graph neural networks
  for rgbd semantic segmentation,'' in \emph{Proceedings of the IEEE
  International Conference on Computer Vision}, 2017, pp. 5199--5208.

\bibitem{bi2019graph}
Y.~Bi, A.~Chadha, A.~Abbas, E.~Bourtsoulatze, and Y.~Andreopoulos,
  ``Graph-based object classification for neuromorphic vision sensing,'' in
  \emph{Proceedings of the IEEE/CVF International Conference on Computer
  Vision}, 2019, pp. 491--501.

\bibitem{garcia2019tactilegcn}
A.~Garcia-Garcia, B.~S. Zapata-Impata, S.~Orts-Escolano, P.~Gil, and
  J.~Garcia-Rodriguez, ``Tactilegcn: A graph convolutional network for
  predicting grasp stability with tactile sensors,'' in \emph{2019
  International Joint Conference on Neural Networks (IJCNN)}.\hskip 1em plus
  0.5em minus 0.4em\relax IEEE, 2019, pp. 1--8.

\bibitem{zhu2018reinforcement}
Y.~Zhu, Z.~Wang, J.~Merel, A.~Rusu, T.~Erez, S.~Cabi, S.~Tunyasuvunakool,
  J.~Kram{\'a}r, R.~Hadsell, N.~de~Freitas \emph{et~al.}, ``Reinforcement and
  imitation learning for diverse visuomotor skills,'' \emph{arXiv preprint
  arXiv:1802.09564}, 2018.

\bibitem{levine2016end}
S.~Levine, C.~Finn, T.~Darrell, and P.~Abbeel, ``End-to-end training of deep
  visuomotor policies,'' \emph{The Journal of Machine Learning Research},
  vol.~17, no.~1, pp. 1334--1373, 2016.

\bibitem{xiang2017posecnn}
Y.~Xiang, T.~Schmidt, V.~Narayanan, and D.~Fox, ``Posecnn: A convolutional
  neural network for 6d object pose estimation in cluttered scenes,''
  \emph{arXiv preprint arXiv:1711.00199}, 2017.

\bibitem{li2018unified}
C.~Li, J.~Bai, and G.~D. Hager, ``A unified framework for multi-view
  multi-class object pose estimation,'' in \emph{Proceedings of the european
  conference on computer vision (eccv)}, 2018, pp. 254--269.

\bibitem{li2019connecting}
Y.~Li, J.-Y. Zhu, R.~Tedrake, and A.~Torralba, ``Connecting touch and vision
  via cross-modal prediction,'' in \emph{Proceedings of the IEEE/CVF Conference
  on Computer Vision and Pattern Recognition}, 2019, pp. 10\,609--10\,618.

\bibitem{scarselli2008graph}
F.~Scarselli, M.~Gori, A.~C. Tsoi, M.~Hagenbuchner, and G.~Monfardini, ``The
  graph neural network model,'' \emph{IEEE transactions on neural networks},
  vol.~20, no.~1, pp. 61--80, 2008.

\bibitem{battaglia2018relational}
P.~W. Battaglia, J.~B. Hamrick, V.~Bapst, A.~Sanchez-Gonzalez, V.~Zambaldi,
  M.~Malinowski, A.~Tacchetti, D.~Raposo, A.~Santoro, R.~Faulkner
  \emph{et~al.}, ``Relational inductive biases, deep learning, and graph
  networks,'' \emph{arXiv preprint arXiv:1806.01261}, 2018.

\bibitem{gilmer2017neural}
J.~Gilmer, S.~S. Schoenholz, P.~F. Riley, O.~Vinyals, and G.~E. Dahl, ``Neural
  message passing for quantum chemistry,'' in \emph{International conference on
  machine learning}.\hskip 1em plus 0.5em minus 0.4em\relax PMLR, 2017, pp.
  1263--1272.

\bibitem{zhou2021pr}
G.~Zhou, H.~Wang, J.~Chen, and D.~Huang, ``Pr-gcn: A deep graph convolutional
  network with point refinement for 6d pose estimation,'' in \emph{Proceedings
  of the IEEE/CVF International Conference on Computer Vision}, 2021, pp.
  2793--2802.

\bibitem{chen2020simple}
M.~Chen, Z.~Wei, Z.~Huang, B.~Ding, and Y.~Li, ``Simple and deep graph
  convolutional networks,'' in \emph{International Conference on Machine
  Learning}.\hskip 1em plus 0.5em minus 0.4em\relax PMLR, 2020, pp. 1725--1735.

\bibitem{sanchez2020learning}
A.~Sanchez-Gonzalez, J.~Godwin, T.~Pfaff, R.~Ying, J.~Leskovec, and
  P.~Battaglia, ``Learning to simulate complex physics with graph networks,''
  in \emph{International Conference on Machine Learning}.\hskip 1em plus 0.5em
  minus 0.4em\relax PMLR, 2020, pp. 8459--8468.

\bibitem{pfaff2020learning}
T.~Pfaff, M.~Fortunato, A.~Sanchez-Gonzalez, and P.~W. Battaglia, ``Learning
  mesh-based simulation with graph networks,'' \emph{arXiv preprint
  arXiv:2010.03409}, 2020.

\bibitem{gao2019graph}
H.~Gao and S.~Ji, ``Graph u-nets,'' in \emph{international conference on
  machine learning}.\hskip 1em plus 0.5em minus 0.4em\relax PMLR, 2019, pp.
  2083--2092.

\bibitem{knyazev2019understanding}
B.~Knyazev, G.~W. Taylor, and M.~Amer, ``Understanding attention and
  generalization in graph neural networks,'' \emph{Advances in neural
  information processing systems}, vol.~32, 2019.

\bibitem{cangea2018towards}
C.~Cangea, P.~Veli{\v{c}}kovi{\'c}, N.~Jovanovi{\'c}, T.~Kipf, and P.~Li{\`o},
  ``Towards sparse hierarchical graph classifiers,'' \emph{arXiv preprint
  arXiv:1811.01287}, 2018.

\bibitem{calli2015ycb}
B.~Calli, A.~Singh, A.~Walsman, S.~Srinivasa, P.~Abbeel, and A.~M. Dollar,
  ``The ycb object and model set: Towards common benchmarks for manipulation
  research,'' in \emph{2015 international conference on advanced robotics
  (ICAR)}.\hskip 1em plus 0.5em minus 0.4em\relax IEEE, 2015, pp. 510--517.

\bibitem{unrealengine}
\BIBentryALTinterwordspacing
{Epic Games}, ``Unreal engine.'' [Online]. Available:
  \url{https://www.unrealengine.com}
\BIBentrySTDinterwordspacing

\bibitem{huynh2009metrics}
D.~Q. Huynh, ``Metrics for 3d rotations: Comparison and analysis,''
  \emph{Journal of Mathematical Imaging and Vision}, vol.~35, no.~2, pp.
  155--164, 2009.

\end{thebibliography}

\end{document}